\documentclass[a4paper]{article}

\usepackage{INTERSPEECH2021}
\usepackage{graphicx}
\usepackage{float} 
\usepackage{subfigure}
\usepackage{bbding}
\usepackage{tablefootnote}

\title{Layer Reduction: Accelerating Conformer-Based Self-Supervised Model via Layer Consistency}
\name{Jinchuan Tian$^1$, Rongzhi Gu$^1$, Helin Wang$^1$, Yuexian Zou$^{1,2}$}
\address{
  $^1$ADSPLAB, School of ECE, Peking University, Shenzhen, China\\
  $^2$Peng Cheng Laboratory, Shenzhen, China}
\email{zouyx@pku.edu.cn}

\begin{document}
\linespread{0.7}
\maketitle

\begin{abstract}
  Transformer-based self-supervised models are trained as feature extractors and have empowered many downstream speech tasks to achieve state-of-the-art performance. However, both the training and inference process of these models may encounter prohibitively high computational cost and large parameter budget. Although Parameter Sharing Strategy (PSS) proposed in ALBERT paves the way for parameter reduction, the computation required remains the same. Interestingly, we found in experiments that distributions of feature embeddings from different Transformer layers are similar when PSS is integrated: a property termed as Layer Consistency (LC) in this paper. Given this similarity of feature distributions, we assume that feature embeddings from different layers would have similar representing power. In this work, Layer Consistency enables us to adopt Transformer-based models in a more efficient manner: the number of Conformer layers in each training iteration could be uniformly sampled and Shallow Layer Inference (SLI) could be applied to reduce the number of layers in inference stage. In experiments, our models are trained with LibriSpeech dataset and then evaluated on both phone classification and Speech Recognition tasks. We experimentally achieve 7.8X parameter reduction, 41.9\% training speedup and 37.7\% inference speedup while maintaining comparable performance with conventional BERT-like self-supervised methods.
  
\end{abstract}

\section{Introduction}
  Unsupervised learning is a learning paradigm aiming at learning effective latent representations without labeling information involved. A special case of unsupervised learning is self-supervised learning \cite{jing2020self}, in which pseudo-labels are obtained from training data itself. Recently, self-supervised learning has shown great potential to empower a wide range of downstream tasks. For example, simSLR \cite{chen2020simple} in Computer Vision (CV) field provides comparable performance with supervised learning in image classification task; word and sentence representations learned from BERT \cite{devlin2018bert}, GPT \cite{radford2018improving} and their followers \cite{liu2019roberta, lan2019albert, chuang2019speechbert} maintain state-of-the-art results in multiple downstream Neural Language Processing (NLP) tasks; Speech representation extractors, like wav2vec \cite{schneider2019wav2vec, baevski2019vq} and TERA \cite{liu2020tera}, provide more informative features and show significant performance improvement in downstream applications like Automatic Speech Recognition (ASR).
  
  There are mainly three self-supervised learning paradigms in speech domain: Autoregressive Predictive Coding (APC) \cite{Chung2019, Chung2020}, Contrastive Predictive Coding (CPC) \cite{oord2018representation, schneider2019wav2vec, baevski2019vq} and Masked Predictive Coding (MPC) \cite{9054458, liu2020tera}, all of which try to encode semantic information (e.g., phonetic information) from contextual speech and output learned features for downstream tasks. Similar to autoregressive language model training in NLP domain, APC tries to predict future frames by encoding previous context in an autoregressive manner. Implementing contrastive learning in speech domain leads to CPC, where the neighbor frames of each frame are forced to have smaller distance compared with randomly sampled frames in latent representation space with a given distance metric. MPC, also known as BERT-like reconstruction method, is a training paradigm where bidirectional speech context is encoded by predicting continuously masked frame blocks. \cite{liu2020tera} is cited as a more comprehensive reference.
  
  In recent practice, models consisting of multiple transformer layers are widely adopted in speech self-supervised learning for better performance in modeling long dependency. Conformer \cite{gulati2020conformer}, as a modified version of Transformer, is subsequently proposed to better exploit local features using convolutional blocks. However, wide applications of these models are impeded by high computational cost and excessed parameter budget. Thus, many methods have been proposed to accelerate Transformer models and to reduce the parameters consumed. Some of these methods are from model compression community, such as quantization \cite{han2016deep}, pruning \cite{NIPS1989_6c9882bb}, low-rank approximation \cite{denton2014exploiting} and knowledge distillations \cite{hinton2015distilling}, while others attempt to design model architectures to obtain better computational and parametric efficiency \cite{wu2020lite, NEURIPS2019_dc960c46}.
  
  ALBERT \cite{lan2019albert}, a lite version of BERT \cite{devlin2018bert}, proposes Parameter Sharing Strategy (PSS) to save parameters. By sharing parameters across all transformer layers, ALBERT achieves multiple times of parameter saving with slight performance degradation. However, computational cost of ALBERT remains the same with conventional BERT model. Next, Audio version of ALBERT \cite{chi2020audio} is subsequently proposed for audio representation learning. 
  
  Provided approaches aforementioned, most Transformer-based models use a deterministic number of layers, and there are few works that try to accelerate Transformers by manipulating this number. Empirically, the computational cost of Transformer-based models is proportional to the number of layers, so reducing this number could speed up these models effectively. In this work, we propose two methods to reduce the equivalent number of layers during training and inference stage respectively:
  \begin{itemize}
  \item Train Conformer-based models with a uniformly distributed number of layers in every iteration. 
  \item Extract features with only first several layers during inference stage (Shallow Layer Inference, SLI).
  \end{itemize}
  The underlying rationality of this proposal could be the Layer Consistency (LC) property provided by PSS, in which the feature distributions across layers maintain a much tighter layout than those Conformer-based models without PSS integrated. To experimentally evaluate the effectiveness of our methods, standard Conformer with 8 layers are trained in MPC style with LibriSpeech 960-hour dataset. The performance of the learned features are evaluated on both phone classification task and ASR task on \textit{train-clean-100} subset. Compared with a standard MPC model, We achieve 7.8X parameter reduction, 41.9\% training speedup and 37.7\% inference speedup while maintaining comparable results in downstream tasks. 

\section{Methodology}
\subsection{Masked Predctive Coding} \label{MPC_method}
\begin{figure}[t]
  \centering
  \includegraphics[width=\linewidth]{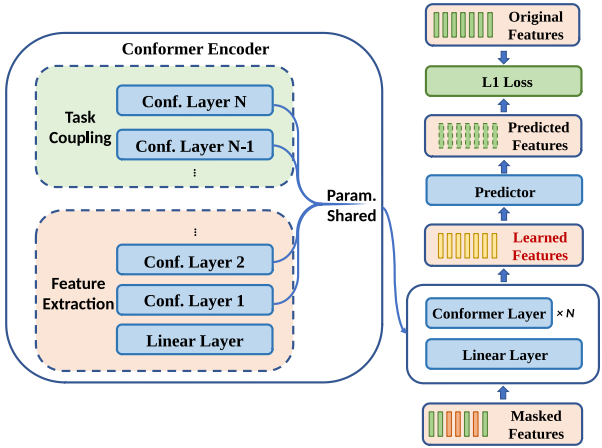}
  \caption{An illustration of Masked Predictive Encoding (MPC) with parameters shared}
  \label{MPC_fig}
\end{figure}
  Following \cite{liu2020tera}, MPC is adopted in this work as the learning paradigm. However, we state that the proposed methods can also be applied in other paradigms like APC and CPC. 
  
  A diagram of MPC is provided in Fig.\ref{MPC_fig}. Hand crafted features (e.g., MFCC, FBANK) are firstly computed from raw speech, and some frames would be randomly masked before being fed into a neural encoder. Subsequently, the learned features would be output by the neural encoder, and the original hand-crafted features are predicted based on these features using a neural predictor. The neural network is optimized through minimizing the frame-level distance (usually L1 distance) between the original features and their predicted counterpart:
  \begin{equation}
      \mathop{\arg\min}\limits_{\mathbf{\theta}} \mathbf{D}(\mathbf{x}, f_{\mathbf{\theta}}(\hat{\mathbf{x}})) = 
      \mathop{\arg\min}\limits_{\mathbf{\theta}} ||\mathbf{x}-f_{\mathbf{\theta}}(\hat{\mathbf{x}})||_{1}
  \end{equation} 
  where $\mathbf{x}$ and $\hat{\mathbf{x}}$ denote the original and masked features respectively, while $f_{\theta}$ represents the combination of encoder and predictor equipped with parameter $\theta$. With the neural encoder being optimized, it learns to encode the bidirectional context and is capable of capturing high-level semantic information. In the inference stage, the learned features could be considered as an alternative of conventional hand-crafted features for downstream tasks for better performance.
  
  Time alteration policy in \cite{liu2020tera} is adopted as the masking mechanism. For each utterance, several continuous frame blocks (7 frames in this work) are masked, which accounts for 15\% of the total frames. The recently proposed Conformer \cite{gulati2020conformer} is chosen as the neural encoder, since it shows greater representing power than conventional Transformer in ASR and has not been comprehensively investigated in self-supervised learning. We use a linear layer as the predictor.
  
\subsection{Parameter Sharing and Layer Consistency}

  Firstly proposed in \cite{lan2019albert}, Parameter Sharing Strategy means using an identical group of parameters for all different Transformer layers. By doing this, the parameter budget is significantly reduced according to the number of the layers in Transformer-Based models. Although maintaining great parameter efficiency improvement, PSS may encounter slight performance degradation as reported in \cite{lan2019albert}.
  
  In this work, we state that PSS also provides another desired property called Layer Consistency (LC). LC means the property that the posterior distributions of feature embeddings across layers are close enough with a given input $\hat{\textbf{x}}$. It can be formulated as:
     \begin{equation}\label{assumption}
       p(\mathbf{e}_i|\hat{\mathbf{x}}) \approx p(\mathbf{e}_j|\hat{\mathbf{x}}) \quad \forall 1\leq i,j \leq N
    \end{equation}
  where $p(\mathbf{e}_i|\hat{\mathbf{x}})$ is the posterior distribution of embedding $\mathbf{e}_i$ from i-th layer.

\begin{figure}
\centering
\subfigure[Conformer-8 Layer + PSS]{
\label{scatter2}
\includegraphics[width=0.22\textwidth]{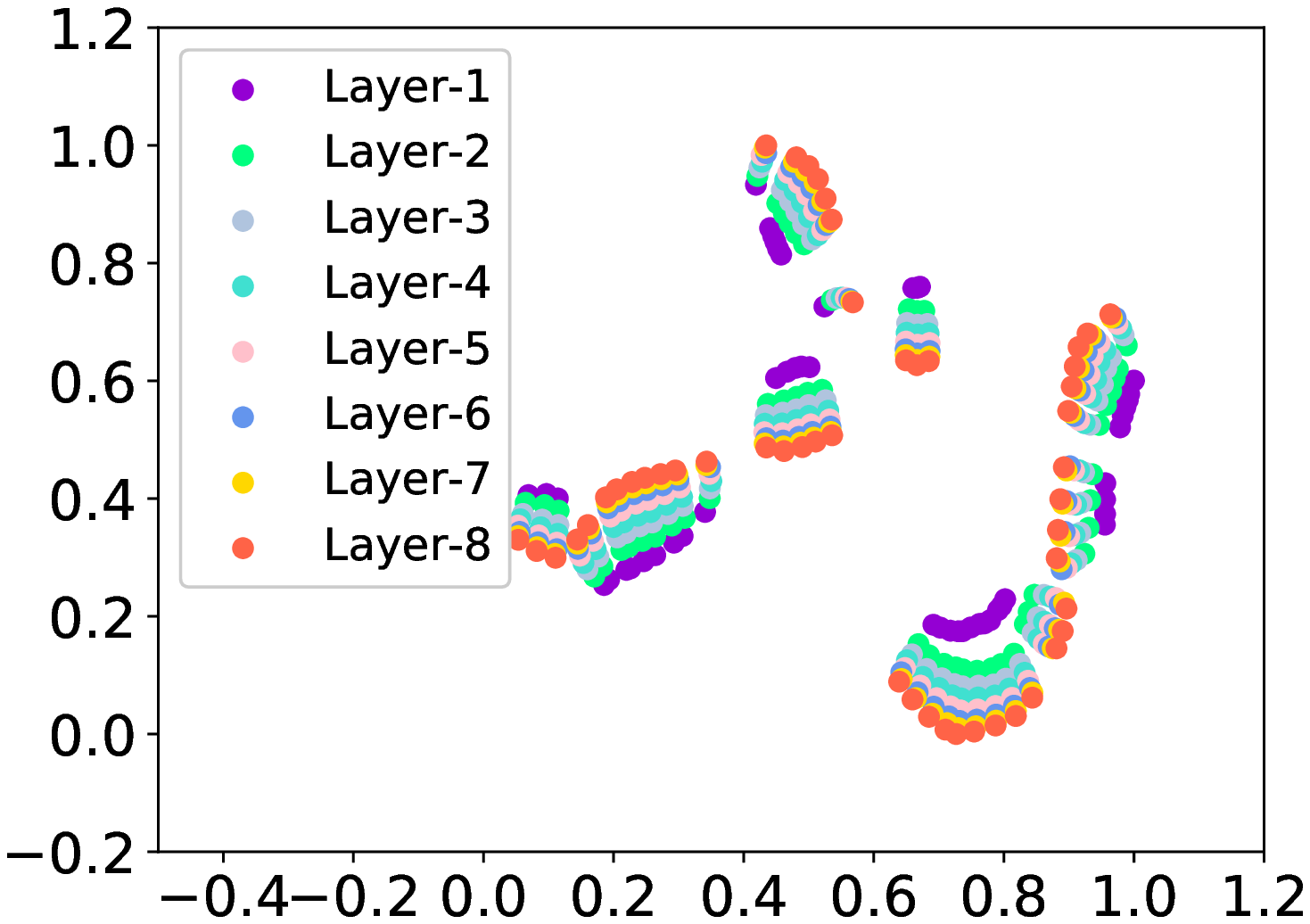}}
\subfigure[Conformer-8 Layer]{
\label{scatter1}
\includegraphics[width=0.22\textwidth]{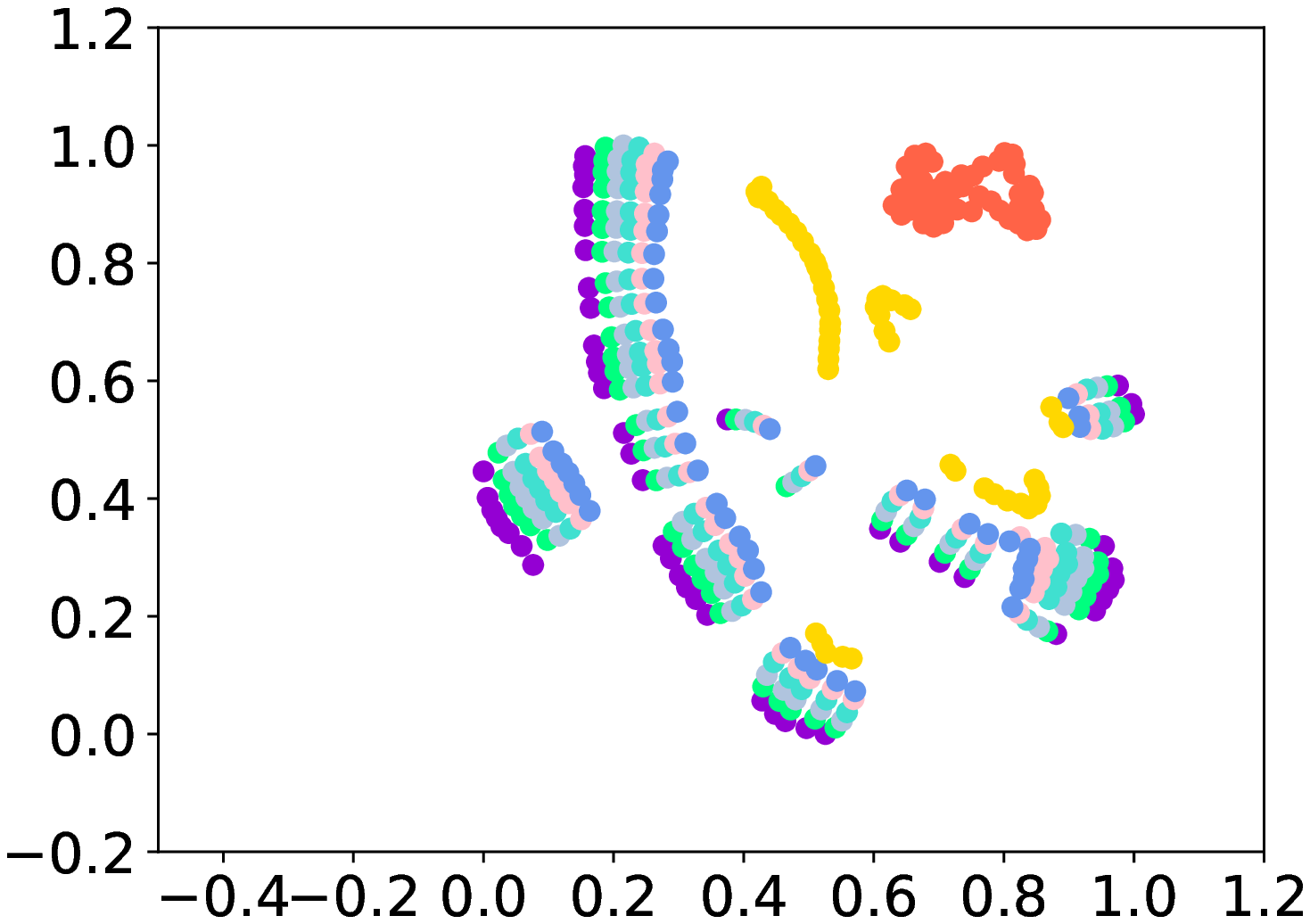}}
\caption{Scatter plots for features across layers. Data obtained after the models have converged. }
\label{fig_scatter}
\end{figure}
  
  To facilitate the understanding of LC, the embedding outputs from different conformer layers are visualized in Fig.\ref{fig_scatter}. With different layer settings, an utterance in \textit{test-clean} subset is encoded layer by layer, and embeddings of 50 continuous frames from each layer are subsequently shown with the feature dimension reduced to 2 using t-SNE\cite{van2008visualizing} technology. Different colors represent different layers. According to these plots, distributions across layers show a compact layout if the PSS is integrated, and a gradual evolving process of these features could be observed by the moving curves of these points. By contrast, clusters for those models without PSS have large gaps between each other and show a more random layout. 
  
\subsection{Layer reduction in training}\label{train_method}
  Based on the statement in equation (\ref{assumption}), posterior distributions of feature embeddings across layers are similar, and then the predictor can make predictions based on embeddings from various layers. Thus, we propose that the number of layers $N$ for each training iteration could be sampled from a uniform distribution if the PSS is integrated:
  \begin{equation}
      N\sim U(L, H)
  \end{equation}
  where $L$ and $H$ are the minimum and maximum number of layers respectively. Given our approach, the expected number of layers would be reduced from $H$ to $\frac{1}{2}(L+H)$, which would, on average, save the computational cost for last $\frac{1}{2}(H-L)$ layers. Our approach maintains not only considerable speedup but also no performance degradation compared with the scenarios using deterministic number of layers as shown in table \ref{tab:main_result}.
  
  Through a more in-depth analysis, the success of our training method could be attributed to Layer Consistency. Assume all vectors are column vectors, and $\frac{\partial \mathbf{J}}{\partial \mathbf{e}_N}$ is the Jacobian Matrix. Given the objective function $\mathbf{J}$, the gradients for parameters $\mathbf{\theta}_i$ from $i$-th layer is back-propagated as:
  
  \begin{equation}
     \left[\frac{\partial \mathbf{J}}{\partial \mathbf{\theta}_i}\right]^{\top} 
     = \left[\frac{\partial \mathbf{J}}{\partial \mathbf{e}_N}\right]^{\top} \cdot 
     \frac{\partial \mathbf{e}_N}{\partial \mathbf{e}_{N-1}} \cdots 
     \frac{\partial \mathbf{e}_i}{\partial \mathbf{\theta}_i}
  \end{equation}
  When using PSS, consider the assumption in equation (\ref{assumption}) and a certain input $\hat{\mathbf{x}}$, for any layer index $i,j$, different number of layers $N, N'$ and shared parameters $\mathbf{\theta}_i = \mathbf{\theta}_j = \mathbf{\theta}$, we have:
  \begin{equation}
      \frac{\partial \mathbf{J}}{\partial \mathbf{e}_N} \approx \frac{\partial \mathbf{J}}{\partial \mathbf{e}_{N'}},
      \quad
      \frac{\partial \mathbf{e}_i}{\partial \mathbf{e}_{i-1}} \approx \mathbf{I},
      \quad
      \frac{\partial \mathbf{e}_i}{\partial \mathbf{\theta}_i} \approx \frac{\partial \mathbf{e}_j}{\partial \mathbf{\theta}_j}
  \end{equation}
  and the gradients for $\theta$ is the sum of gradients from all layers:
  \begin{equation}
  \begin{aligned}
      \left[\frac{\partial \mathbf{J}}{\partial \mathbf{\theta}}\right]^{\top}
      & = \sum_{i=1}^N \left[\frac{\partial \mathbf{J}}{\partial \mathbf{\theta_i}}\right]^{\top}  \\
      & = \sum_{i=1}^N \left[\frac{\partial \mathbf{J}}{\partial \mathbf{e}_N}\right]^{\top} \cdot 
          \frac{\partial \mathbf{e}_N}{\partial \mathbf{e}_{N-1}} \cdots 
          \frac{\partial \mathbf{e}_i}{\partial \mathbf{\theta}_i}\\
      & \approx \sum_{i=1}^N \left[\frac{\partial \mathbf{J}}{\partial \mathbf{e}_N}\right]^{\top} \cdot 
        \frac{\partial \mathbf{e}_i}{\partial \mathbf{\theta}_i} \\
      & = N \cdot 
         \left[\frac{\partial \mathbf{J}}{\partial \mathbf{e}_N}\right]^{\top} \cdot 
         \frac{\partial \mathbf{e}_i}{\partial \mathbf{\theta}_i}. \\
  \end{aligned}
  \end{equation}
  Thus, we state that varying $N$ in each training iteration could be approximated by a scaling operation on gradients with $N$ being the factor (equivalent to scaling on learning rate) when the optimizer asymptotically approaches the local minimum. By contrast, if PSS is not involved, model performance could be compromised by the varying $N$ due to the mismatch on updating frequency between shallow layer parameters and deep layer parameters.

\subsection{Layer reduction in inference} \label{infer_method}
  In self-supervised learning tasks equipped with Transformer-based models, not all of the layers serve the purpose of encoding context and capturing high-level semantic information. Instead, as illustrated in the left part of Fig.\ref{MPC_fig}, last several layers may try to translate hidden representation among layers to a space where the original features are easier to predict from, which is an unwanted coupling phenomenon. Given this observation, we state that computation of these coupling layers would bring no further benefits in downstream tasks, which is experimentally validated in section \ref{layer_performance}.
  
  Considering the coupling phenomenon, Shallow Layer Inference (SLI) is proposed to speed up inference by reducing number of layers to be computed. Given a Transformer encoder with multiple layers, not all of these layers are necessarily computed during inference no matter it is trained with deterministic or random number of layers. Instead, only the first several layers are computed during inference. SLI($M$) stands for inference only with first $M$ layers.
  
\section{Experiments}
\subsection{Experimental setup}\label{setup}
  Time alteration in TERA \cite{liu2020tera} is adopted as an implementation of MPC learning paradigm. The number of Conformer layers is 8 unless other specified. The model is trained with LibriSpeech \cite{panayotov2015librispeech} 960-hour dataset. 80-dim FBANK spectrum plus 3-dim pitch features are combined as the input hand-crafted features. In each Conformer layer, we use 512 hidden units, 4 attention heads and 0.1 attention dropout rate in attention block, while the hidden size of feed forward block and filter number of convolutional block are designed as 2048 and 15 respectively. Each Conformer layer consumes 4.3M parameters. Adam optimizer \cite{kingma2015adam} with noam learning rate schedule \cite{NIPS2017_3f5ee243} is used, and the warm up steps is set to 8K. The model are trained on 4 GPUs for 1M iterations (30 epochs) with mini-batch size of 8, and models with best validation loss are kept by evaluating \textit{dev-clean} and \textit{dev-other} set. No ensemble method is used. Code of this stage is revised from Espnet \cite{hayashi2020espnet}.
  
  The effectiveness of learned features are evaluated on two downstream task: phone classification and ASR. In phone classification task, linear separability of learned features are evaluated using \textit{train-clean-100} subset of Librispeech following the split of CPC \cite{oord2018representation}, and the alignment labels are obtained by \textit{tri6b} model in Kaldi \cite{povey2011kaldi} recipe. In ASR task, a standard DNN-HMM model is built with \textit{train-clean-100} using the learned features, then the Word Error Rates (WERs) with and without rescoring are reported. A 5-layer TDNN with hidden size 1280 and strides of context \{1, 4, 7, 10, 1\} for each layer is adopted as the acoustic model and is trained for 4 epochs. A 5-gram language model provided by Librispeech is used for rescoring. Code of ASR evaluation is implemented in Kaldi. We adopt DNN-HMM models to avoid the interference of neural language modeling.

\subsection{Acceleration results}\label{speed_result}
\begin{table*}
  \caption{Results of different models with multiple metrics. Parameter budget, phone classification accuracy, WER with/without rescoring, time consumed in each epoch and real time factor are reported in order}
  \label{tab:main_result}
  \centering
  \begin{tabular}{|l|c|c|c|c|c|c|}
    \toprule
    \textbf{Model}  &\textbf{Param.} &\textbf{Phone Class.}  &\textbf{WER}  &\textbf{Rescore}  &\textbf{GPU$\cdot$h / Epoch\tablefootnote{Test on single GPU to avoid latency of gradient exchange}}  &\textbf{RTF(CPU)} \\
    \midrule
    MFCC                                         & -     & 38.7   & 10.67 & 7.46 & - & -    \\
    CPC\cite{oord2018representation} - 2018      & -     & 64.6   & -     & -    & - & -    \\
    liGRU+Mockingjay\cite{9054458}   - 2020      & 21.3M & 67.0   & 8.31  & 5.99 & - & -    \\
    liGRU+TERA-medium\cite{liu2020tera}  -2020    & 42.6M & 67.3   & 8.37  & 6.05 & - & -    \\
    \midrule
    Exp.1 Conformer-8L                             & 33.7M & 72.70           &\textbf{7.80}   & \textbf{5.77} & 8.47          & 0.506          \\
    Exp.2 Conformer-8L + SLI(5)                    & 33.7M & \textbf{72.92}  &7.90            & 5.83          & 8.47          & 0.315          \\
    \midrule
    Exp.3 Conformer-5L + PSS                       & 4.3M  & 73.07           & 8.48           & 6.43          & 4.93          & \textbf{0.311} \\
    Exp.4 Conformer-6L + PSS                       & 4.3M  & 73.27           & 8.71           & 6.36          & 5.64          & 0.372          \\
    Exp.5 Conformer-8L + PSS                       & 4.3M  & 74.17           & 8.26           & \textbf{6.12} & 7.09          & 0.497          \\
    Exp.6 Conformer-$U(2,8)$L + PSS                & 4.3M  & 73.44           & 8.34           & 6.21          & \textbf{4.92} & 0.502          \\
    Exp.7 Conformer-$U(2,8)$L + PSS + SLI(5)       & 4.3M  & 72.98           & 8.30           & \textbf{6.12} & \textbf{4.92} & 0.315          \\
    Exp.8 Conformer-$U(4,8)$L + PSS                & 4.3M  & 74.32           & 8.30           & 6.15          & 5.66          & 0.498          \\
    Exp.9 Conformer-$U(4,8)$L + PSS + SLI(5)       & 4.3M  & \textbf{74.41}  & \textbf{8.25}  & 6.15          & 5.66          & 0.321          \\
    \bottomrule
  \end{tabular}
\end{table*}
  In table \ref{tab:main_result}, acceleration performance of our methods are reported. Other metrics, like parameter budget, phone classification accuracy and WER with/without rescoring, are also listed as reference. 
  Speed performance with different settings are firstly compared to show the advantages of the proposed methods. Compared with Exp.1, Exp.6 and Exp.7 achieve 41.9\% training speedup for two reasons: (1) the expected number of layers is reduced from 8 to 5; (2) less parameters to be updated due to PSS. Subsequently, Exp.7 maintains 37.7\% inference speedup compared with Exp.1, since only first 5 layers are used during inference. Finally, experiments with PSS achieve 7.8X parameter reduction than conventional approach (Exp.1).
  
  Secondly, results for downstream tasks are compared. In terms of phone classification, all 8-layer conformers with PSS achieve better linear separability than those without PSS, and the best result is observed with all our methods integrated (Exp.9). For ASR task, the application of PSS results in around 0.5\% absolute WER degradation compared with conventional approach (Exp.1), which is still an unsolved problem as reported in original ALBERT \cite{lan2019albert}. However, compared with Exp.5, both our methods achieve much close performance (Exp.6-9) while maintaining the speedups. Also, given similar computational resource, a performance gap is observed between Exp.3,4 and Exp.7,9, which suggests the necessity of this work.
  
  We also compare our work with other previous works. Though most of previous works adopted Transformers as the backbone, the adoption of Conformer leads to a performance gain on both downstream tasks. Consider the gain caused by Conformer and the degradation resulted from PSS, we achieve comparable performance in downstream tasks and save more than 1/3 of computation in both training and inference stages.

\subsection{Comparison across layers}
\subsubsection{Performance across layers}\label{layer_performance}

\begin{table}[t]
  \renewcommand\arraystretch{1}
  \scriptsize
  \caption{Phone classification and ASR results across layers. Phone accuracy and WERs with / without rescoring are reported with varying inference layers and parameter policy}
  \label{tab:layer_result}
  \centering
  \setlength{\tabcolsep}{4mm}{
  \begin{tabular}{|c|c|c|c|c|}
    \toprule
    Inf. Layer & PSS & Ph.Cls & WER & Rescore  \\
    \midrule
    \multicolumn{5}{|l|}{Conformer-8L} \\
    \midrule
    5       & \XSolidBrush & 72.92          & 7.90         & 5.83 \\
    6       & \XSolidBrush & 75.71          &\textbf{7.80} & 5.86 \\
    7       & \XSolidBrush & \textbf{76.33} & 7.81         & \textbf{5.74} \\
    8       & \XSolidBrush & 72.70          &\textbf{7.80} & 5.77 \\
    \midrule
    5       & \Checkmark  & 74.76           & 8.18         & \textbf{6.09} \\
    6       & \Checkmark  & \textbf{74.87}  & 8.26         & 6.12 \\
    7       & \Checkmark  & 74.59           &\textbf{8.17} & 6.14 \\
    8       & \Checkmark  & 74.17           & 8.26         & 6.12 \\
    \midrule
    \multicolumn{5}{|l|}{Conformer-$U(2,8)$L} \\
    \midrule
    5       & \XSolidBrush & \textbf{72.95} & 8.92          & 6.59             \\
    6       & \XSolidBrush & 72.92          & \textbf{8.82} & \textbf{6.52}    \\
    7       & \XSolidBrush & 72.94          & 8.93          & 6.59             \\
    8       & \XSolidBrush & \textbf{72.95} & 8.91          & 6.56             \\
    \midrule
    5       & \Checkmark  & 73.44           & 8.30          & \textbf{6.12}    \\
    6       & \Checkmark  & \textbf{73.48}  & 8.30          & 6.15             \\
    7       & \Checkmark  & 73.29           & \textbf{8.29} & 6.13             \\
    8       & \Checkmark  & 72.98           & 8.34          & 6.21             \\
    \bottomrule
  \end{tabular}}
\end{table}
  Accuracy performance with different numbers of inference layers are reported in table \ref{tab:layer_result}, where the results of phone classification and WER with or without rescoring are listed. Results of Conformer-$U(4,8)$L are consistent with those of Conformer-$U(2,8)$L. They are not reported due to the limited space.
  
  For both downstream tasks, nearly no degradation is observed across layers, which verifies the superiority of SLI. For phone classification task, \textit{Conformer-8} without PSS achieves the highest accuracy in the 7-th layer with much fluctuation across layers. However, models with PSS and our methods integrated (\textit{Conformer-U(2,8)} consistently achieve considerable results. Next, results of ASR task show similar stability in all cases, and these results further validate the rationality of our SLI mechanism. We argue that feature embeddings from last several layers maintain similar representing power due to the coupling phenomenon. Finally, without PSS, there is a big performance gap between \textit{Conformer-U(4,8)L} and \textit{Conformer-8L}, which is compatible with our theory in section \ref{train_method}.

\subsubsection{Layer transitions}

\begin{figure}
\centering  
\subfigure[L2 Distance]{
\label{Fig.sub.1}
\includegraphics[width=0.22\textwidth]{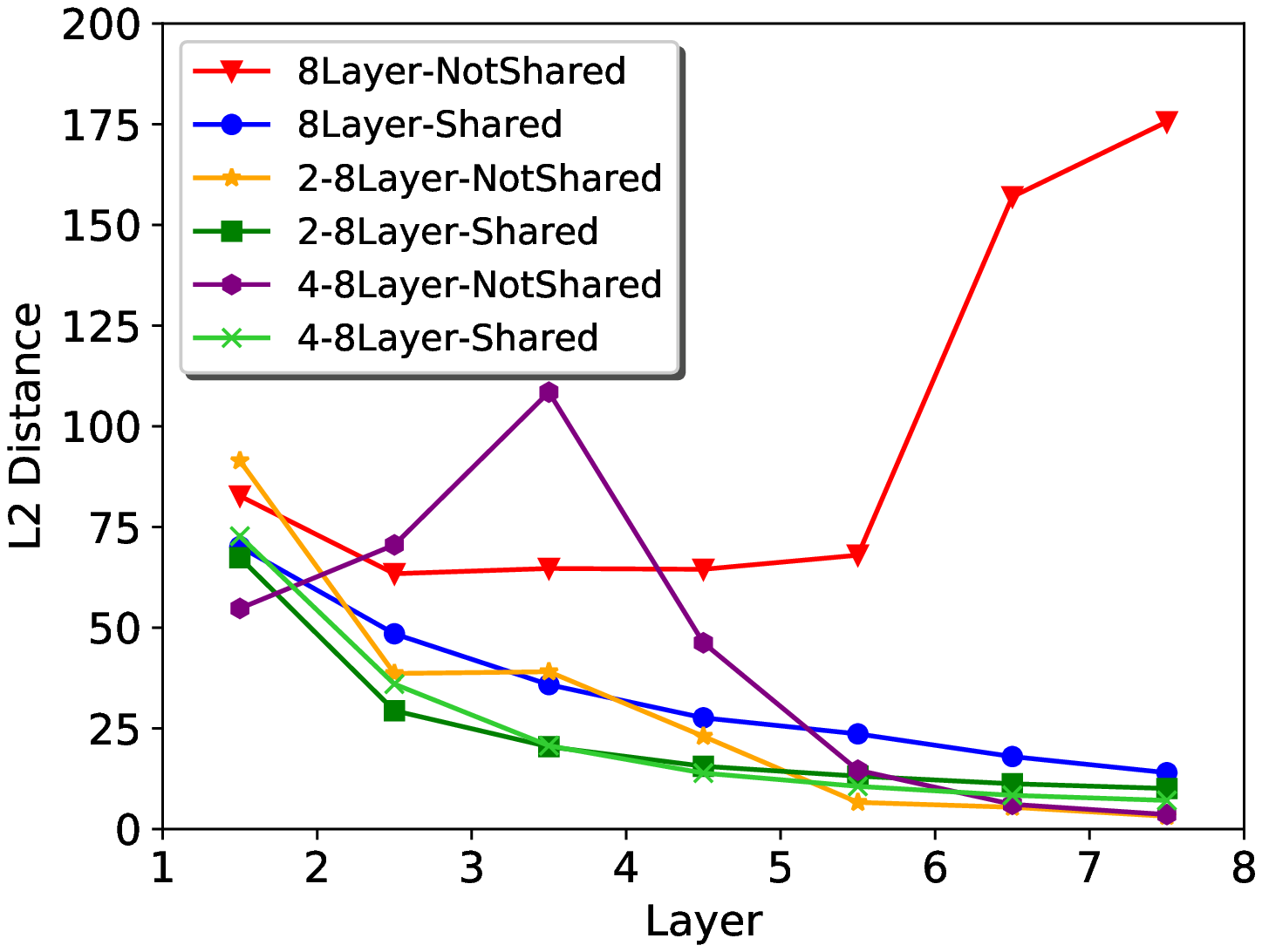}}
\subfigure[Cosine Similarity]{
\label{Fig.sub.2}
\includegraphics[width=0.22\textwidth]{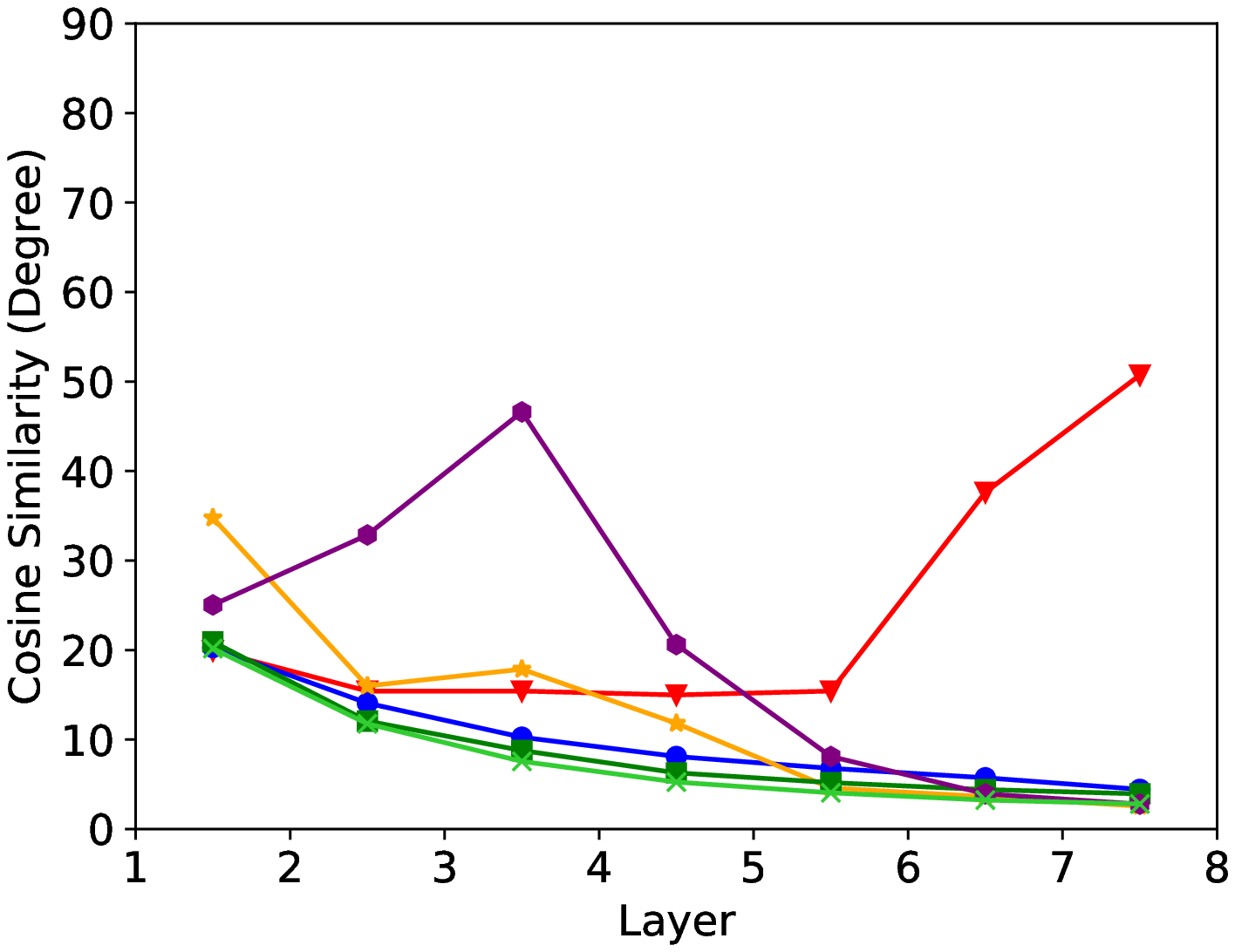}}
\caption{Transition results across layers}
\label{fig_transition}
\end{figure}

  Layer consistency is further demonstrated by comparing the transition (L2 distance and cosine similarity) between each two continuous layers (see Fig.\ref{fig_transition}). Models without PSS show fierce transformation from layer to layer, while feature embeddings in \textit{8Layer-Shared}, \textit{2-8Layer-Shared} and \textit{4-8Layer-Shared} have decreasingly small distances from low-level features to high-level features. Layer Consistency caused by PSS also delivers us the possibility to adopt features from different layers in an adaptive manner, and a trade-off between performance and speed could be achieved with adjustable settings. We leave this for our future work.
  
\section{Conclusion}
  
  This work explores the Layer Consistency of Conformer-based self-supervised models with Parameter Sharing Strategy integrated to accelerating their training and inference process. By training Conformer-based models with a uniformly distributed number of layers in each iteration, we achieve 41.9\% speedup during training. Also, application of Shallow Layer Inference during inference leads to 37.7\% reduction in computational complexity. Comprehensive experiments have been conducted on two downstream tasks and nearly no performance degradation is observed compared with other advanced baselines even with 7.8X parameter reduction. The speedups and parameter reduction achieved in this work effectively support our statement that embeddings from different layers have similar representing power when using Parameter Sharing Strategy.
  

\bibliographystyle{IEEEtran}

\bibliography{det_version}
\end{document}